\documentclass{article}

\PassOptionsToPackage{numbers, compress}{natbib}

\usepackage[preprint]{neurips_2025}
\usepackage{xcolor}
\usepackage{soul}

\sethlcolor{white}
\usepackage{longtable}
\usepackage[utf8]{inputenc} 
\usepackage[T1]{fontenc}    
\usepackage{hyperref}       
\usepackage{url}            
\usepackage{booktabs}       
\usepackage{amsfonts}       
\usepackage{nicefrac}       
\usepackage{microtype}      
\usepackage{xcolor}         
\usepackage{pifont}
\usepackage{enumitem}
\usepackage{amsmath}
\usepackage{algorithm}
\usepackage{algpseudocode}
\usepackage{subcaption}
\usepackage{xspace}
\usepackage{cleveref}
\definecolor{brightblue}{RGB}{0, 127, 255}
\definecolor{brightgreen}{RGB}{0, 204, 0}
\usepackage{graphicx}
\usepackage{multirow, makecell}

\usepackage{colortbl}  

\usepackage{lineno}
\usepackage{caption}  
\definecolor{darkblue}{rgb}{0, 0, 0.5}
\hypersetup{colorlinks=true, citecolor=darkblue, linkcolor=magenta, urlcolor=magenta}

\usepackage{graphicx}
 \usepackage{listings}

\definecolor{codebackground}{rgb}{0.95,0.95,0.95}
\definecolor{codekeyword}{rgb}{0.13,0.29,0.53}
\definecolor{codestring}{rgb}{0.31,0.60,0.02}
\definecolor{codecomment}{rgb}{0.5,0.5,0.5}

\newcommand{\ourmethod}{\textsc{FlowReasoner}\xspace}

\title{FlowReasoner: Reinforcing Query-Level Meta-Agents}

\renewcommand\footnotemark{}
\author{%
  Hongcheng Gao$^{*1,2}$, Yue Liu$^{*3}$, Yufei He$^{3}$, Longxu Dou$^{1}$, Chao Du$^{1}$ \\ \textbf{Zhijie Deng}$^{4}$,
  \textbf{Bryan Hooi}$^{3}$, \textbf{Min Lin}$^{1}$, \textbf{Tianyu Pang}$^{\dagger1}$
  \thanks{\llap{$^*$}Equal contribution. $\dagger$ Corresponding author.}\\
  $^{1}$Sea AI Lab, Singapore $^{2}$University of Chinese Academy of Sciences \\
  $^{3}$National University of Singapore $^{4}$Shanghai Jiao Tong University \\
  \footnotesize{\texttt{gaohongcheng23@mails.ucas.ac.cn, yliu@u.nus.edu}}
}

\begin{document}

\maketitle

\begin{abstract}


This paper proposes a query-level meta-agent named \ourmethod to automate the design of query-level multi-agent systems, i.e., \emph{one system per user query}. 
Our core idea is to incentivize a reasoning-based meta-agent via external execution feedback.
Concretely, by distilling DeepSeek R1, we first endow the basic reasoning ability regarding the generation of multi-agent systems to \ourmethod. 
Then, we further enhance it via reinforcement learning (RL) with external execution feedback.
A multi-purpose reward is designed to guide the RL training from aspects of performance, complexity, and efficiency. 
In this manner, \ourmethod is enabled to generate a personalized multi-agent system for each user query via deliberative reasoning. 
Experiments on both engineering and competition code benchmarks demonstrate the superiority of \ourmethod. 
Remarkably, it surpasses o1-mini by $\mathbf{10.52}$\% accuracy across three benchmarks. 
The code is available at \url{https://github.com/sail-sg/FlowReasoner}.




\end{abstract}

\section{Introduction}
Large language models (LLMs)~\citep{gpt_4,gemini1_5,claude,qwen2_5,deepseek_v3} have exhibited remarkable power in various meaningful yet challenging domains, like chatbots~\citep{chatgpt}, code~\citep{devin}, math~\citep{o1}, robotics~\citep{openvla}, etc. LLM-based multi-agent systems~\citep{metagpt,autogen,CAMEL}, which are characterized by planning, reasoning, tool use, and memory, become the foundation of these LLM-driven applications.\footnote{This paper defines a \emph{multi-agent system} as a system consisting of multiple agents operating under a \emph{workflow}.} While effective, most of them are manually designed, increasing human resource costs and limiting scalability.



To mitigate this challenge, early automatic methods are proposed to optimize the prompts~\citep{yuksekgonul2024textgrad,Dspy,zhou2024self,yang2023large} or hyper-parameters~\citep{saad2024archon}. But they still rely on the fixed workflow of the multi-agent system, which requires human effort to manually design workflows for each new scenario. From this motivation, various graph-based methods~\citep{GPTSwarm,DyLAN,g_designer,feng2025heterogeneous} formulate the workflows as graphs or networks and automate the workflow designs. However, the structural complexity of graphs limits their scalability~\citep{ADAS}. To overcome this limitation, state-of-the-art methods represent the multi-agent systems as programming codes~\citep{ADAS} and prompt a performant LLM, e.g., GPT-4o, as a \emph{meta-agent} to optimize workflows via complex search algorithms on carefully designed search sets~\citep{AFLOW,Agentsquare,MaAS}.

\begin{figure*}[t]
\centering
\vspace{-0.cm}
\includegraphics[width=0.95\linewidth]{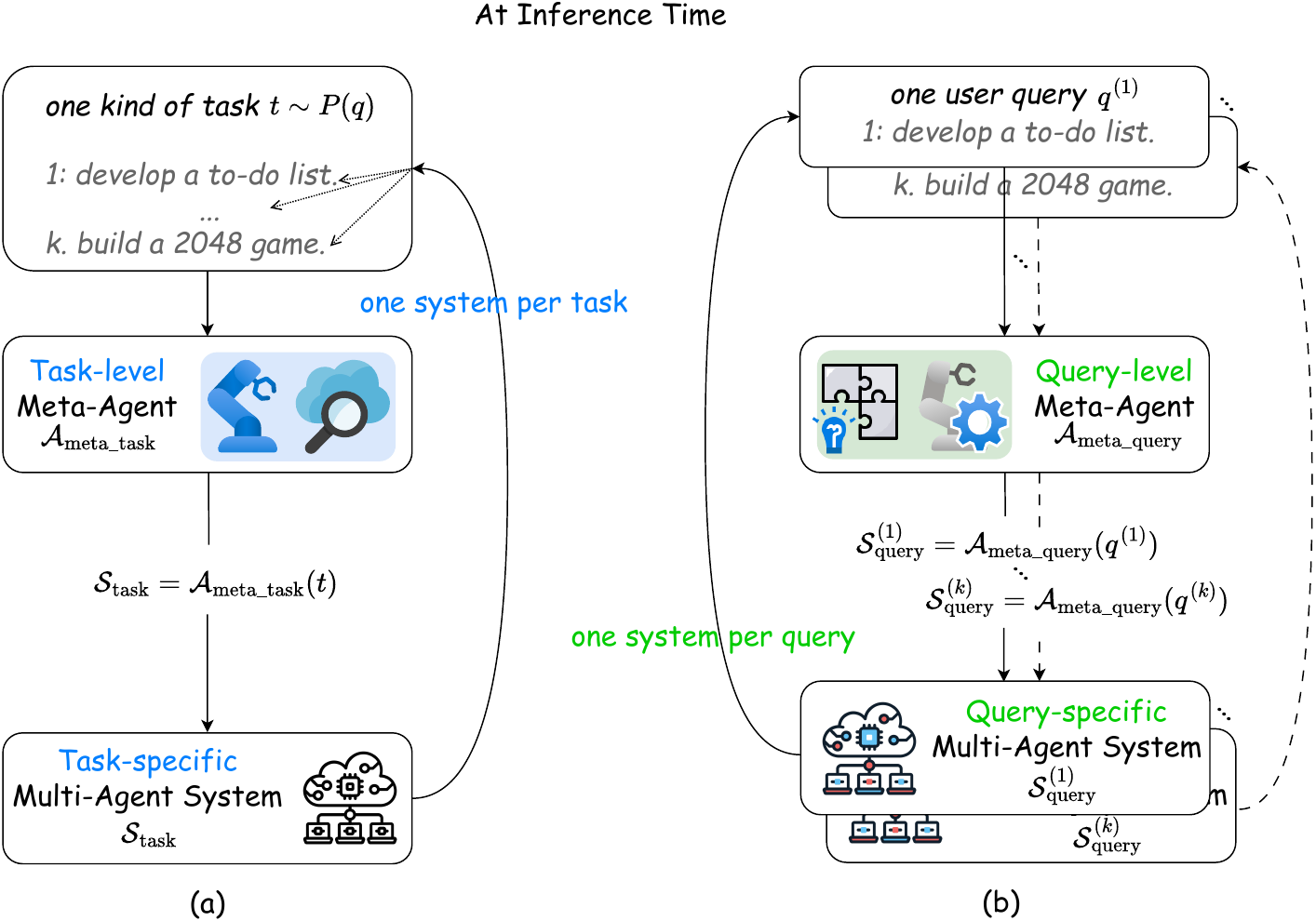}
   \caption{\textbf{Task-Level Meta-Agents vs. Query-Level Meta-Agents at Inference Time.} $q$ denotes a user query, e.g., \textit{build a 2048 game}. $t\sim P(q)$ denotes one kind of task, e.g., \textit{code generation task}, which is a distribution of user queries. Given $t$, previous task-level meta-agent $\mathcal{A}_{\text{meta\_task}}$ aims to search a task-specific multi-agent system $\mathcal{S}_{\text{task}}$ to solve all queries sampled from $t$, i.e., {\color{brightblue}one system per task}. Differently, given one user query $q^{(i)}$, our query-level meta-agent $\mathcal{A}_{\text{meta\_query}}$ conducts reasoning and output a query-specific multi-agent system $\mathcal{S}_{\text{query}}^{(i)}$ for $q^{(i)}$, i.e., {\color{brightgreen}one system per query}.} 
   \vspace{0.3cm}
    \label{fig:drawbacks}
\end{figure*}

These previous methods focus on \textbf{task-level meta-agents}, generating merely a single \textbf{task-specific multi-agent system} that applies to one kind of task, e.g., \textit{code generation task}, as in Figure \ref{fig:drawbacks} (a). 
However, for individual user queries, these one-size-fits-all systems lack the capability for automatic adaptation. 
To enhance the adaptability of multi-agent systems for individual user queries, this paper aims to design a \textbf{query-level meta-agent} to generate a \textbf{query-specific multi-agent system} for each user query, e.g., \textit{build a 2048 game}, as shown in Figure \ref{fig:drawbacks} (b).


We first identify that the success of task-level meta-agents largely depends on carefully designed search sets, as they rely on complex search algorithms. 
However, such search sets are unavailable in the setting of query-specific multi-agent systems.
To address this issue, instead of relying on search algorithms, we propose to integrate \textbf{external execution feedback} of the generated multi-agent system, based on which a \textbf{reasoning-driven meta-agent} is leveraged to polish the system.

We dub such a meta-agent as \ourmethod. 
We first synthesize thousands of warm-up SFT data using DeepSeek R1-671B~\citep{deepseek_r1} as the meta-agent to generate multi-agent systems and process user queries individually. 
These synthetic data are then used to finetune DeepSeek-R1-Distill-Qwen-7B, enabling basic reasoning for multi-agent system generation. 
Furthermore, we enhance its reasoning capabilities for generating novel query-level multi-agent systems through reinforcement learning (RL), incorporating external execution feedback. 
A \textbf{multi-purpose reward} is designed to guide RL training, focusing on \textit{performance}, \textit{complexity}, and \textit{efficiency}.
During inference, \ourmethod leverages deliberative reasoning to generate a novel query-level multi-agent system for each user query, achieving \textit{one system per user query}. The main contributions are summarized as follows:
\begin{enumerate}[label=\textbullet, leftmargin=0.4cm, itemsep=0.2em, parsep=0.2em, topsep=0.em]
    \item We propose a query-level meta-agent termed \ourmethod to automate the designs of query-level multi-agent systems, improving their adaptability in real-world scenarios.
    \item We train \ourmethod to reason from external execution feedback via RL, guided by a multi-purpose reward considering performance, complexity, and efficiency.
    \item We demonstrate the superiority of \ourmethod via extensive experiments and open-source it.  
\end{enumerate}

\vspace{-0.15cm}
\section{Related Work}
\vspace{-0.1cm}
\textbf{LLM-Based Multi-Agent Systems.}
LLM-based multi-agent systems \citep{survey,llm-debate,agentverse,park2023generative,yichong_nips} serve as the foundation of various LLM-powered real-world applications, e.g., code intelligence \citep{devin}, computer use \citep{computer_use}, and deep research \citep{deep_research}. LLM-based agents are equipped with planning capabilities, database access, and tool function invocation. These agents collaborate within a multi-agent system, leading to promising performance.
Most multi-agent systems are manually designed, increasing the costs of human resources and limiting the scalability. To address this issue, researchers propose automation methods to automate the design of multi-agent systems. Early methods \citep{yang2023large,Dspy,zhou2024self} are proposed to automate the agentic designs via optimizing prompts \citep{yuksekgonul2024textgrad} or hyper-parameters \citep{saad2024archon}. For example, \citep{chen2023autoagents,yuan2024evoagent,fernando2023promptbreeder} adopt the evolution algorithms to automate the agent profiling. 


Although effective, they merely optimize the agents but keep the workflow of the multi-agent system fixed, which still requires human effort to manually design for each new scenario. To address this problem, several methods \citep{li2024autoflow,zhou2024symbolic,zhang2024offline,Agentsquare,zhuge2023mindstorms} are proposed to automate the design of the entire agentic workflow. For example, researchers \citep{GPTSwarm,Dspy,DyLAN,g_designer,feng2025heterogeneous} formulate the workflows as graphs or networks and then optimize the connections between nodes. To improve the efficiency, ADAS \citep{ADAS} proposes to use programming codes to represent both agents and workflows. It also introduces a meta-agent to generate these workflows and presents the meta-agent search to optimize the designs of multi-agent systems. At the same time, AFLOW \citep{AFLOW} also adopts the code representation but proposes Monte Carlo Tree Search (MCTS) to optimize it. In addition, MaAS \citep{MaAS} presents the agentic supernet and then conducts the multi-agent architecture search. Differently, ScoreFlow \citep{wang2025scoreflow} trains a workflow generator to generate better workflows via direct preference optimization (DPO) \citep{dpo}.\looseness=-1

This paper argues that the previous methods are task-level meta-agents. As shown in Figure  \ref{fig:drawbacks} (a), they merely generate a single task-specific multi-agent system for solving one kind of task. However, these one-size-fits-all systems are rigid and unable to automatically adapt or customize to individual user queries within a task. From this motivation, we aim to propose a query-level meta-agent to generate a query-specific multi-agent system for each user query as shown in Figure \ref{fig:drawbacks} (b).



\textbf{Reasoning in LLMs.}
The ability to reason is essential for LLMs, enabling them to emulate human thinking patterns. Pioneering work \citep{CoT, kojima2022large} has facilitated this by prompting LLMs to think step by step. Beyond this approach, reasoning capabilities are further enhanced through frameworks such as self-correction \citep{self_correct_1}, self-critique \citep{self_critique}, debate \citep{debate_1, llm-debate}, and plan-and-solve \citep{wang2023plan}. Additionally, efforts like \citep{latent_space, pause_token} seek to transition LLMs’ reasoning processes into the latent space. OpenAI has advanced reasoning in LLMs by developing the o1 model, demonstrating the potential for improvement through test-time scaling. Inspired by this, models such as QwQ \citep{qwq}, QvQ \citep{qvq}, DeepSeek \citep{deepseek_r1}, and Kimi \citep{kimi_k1_5} have followed suit, developing o1-like reasoning architectures. Moreover, OpenAI's o3 model has been announced to achieve promising results on the ARG-AGI benchmark \citep{ARC_AGI}. LLMs progressively shift from intuitive processing (System 1) to deliberative reasoning (System 2) \citep{li2025system}. Besides, researchers demonstrate that reasoning can improve safety \citep{liu2025guardreasoner} and alleviate hallucination \citep{gao2025exploring}. However, \cite{overthinking} examines the overthinking problem observed in o1-like models. To alleviate this problem, token efficiency methods \citep{token_efficiency_survey} are proposed to reduce the token costs while maintaining the reasoning quality. This paper develops an o1-like reasoning model to serve as a query-level meta-agent, getting rid of complex search algorithms and the carefully designed search set.

\section{Problem Definition}

We denote a user \textbf{query} as $q$, e.g., \textit{build a 2048 game}. Then a user \textbf{task} is defined as a distribution of user queries, denoted as $t=P(q)$, e.g., \textit{code generation task}. A multi-agent system is denoted as $\mathcal{S}=\{\mathcal{A},\mathcal{W}\}$, where $\mathcal{A}=\{\mathcal{A}_1,...,\mathcal{A}_n\}$ denotes the agents in the system and $\mathcal{W}$ is the \emph{workflow} of collaboration among the agents.

\begin{figure*}[t]
\centering
\includegraphics[width=\linewidth]{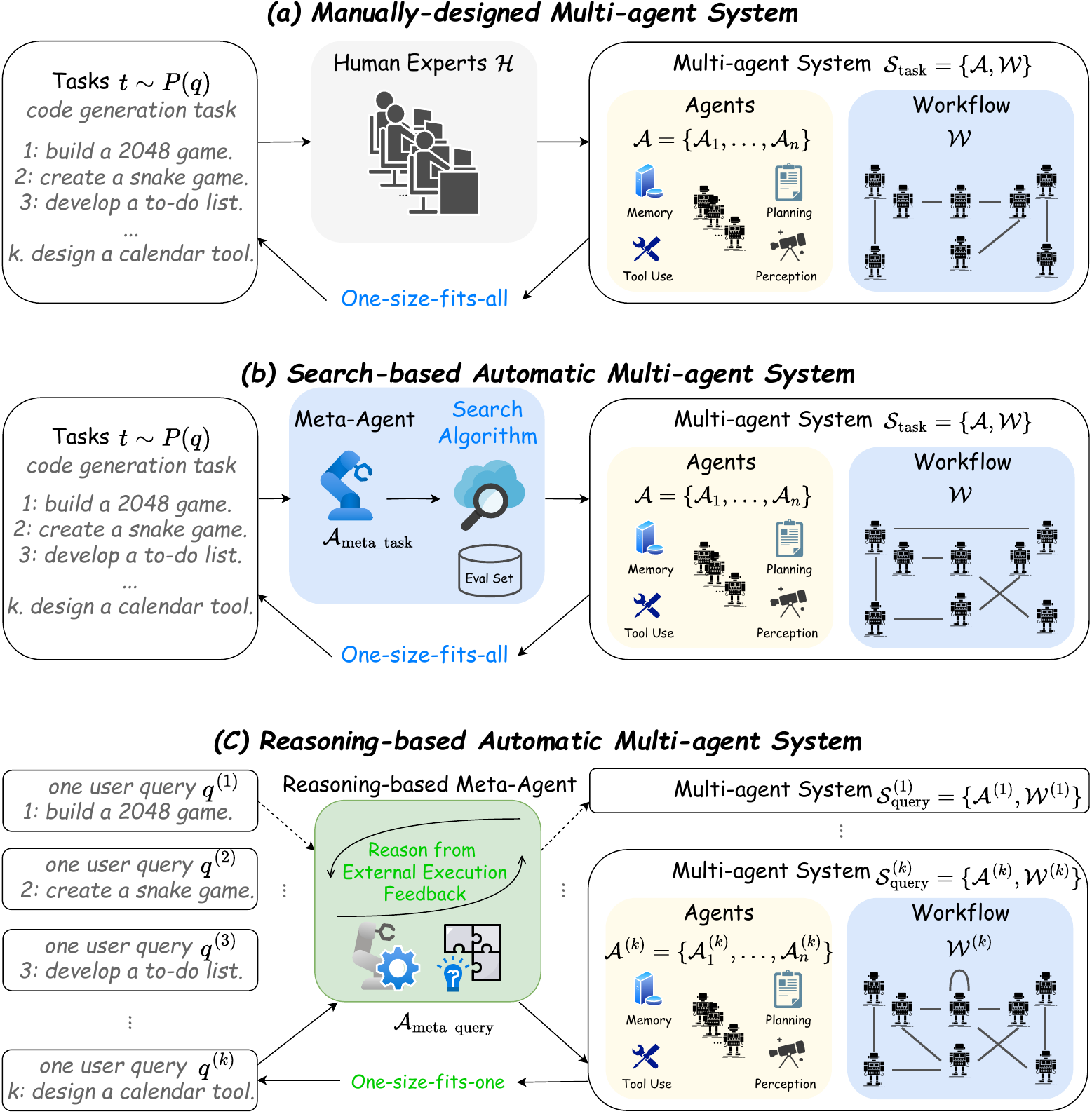}
   \caption{\textbf{Architectural Comparison of Three Multi-Agent Systems}. (a) Manually-designed Multi-agent System, (b) Search-based Automatic Multi-agent System, and (c) Reasoning-based Automatic Multi-agent System.} 
    \label{fig:diff}
    \vspace{0.3cm}
\end{figure*}

As shown in Figure \ref{fig:diff} (a), in traditional multi-agent systems \citep{metagpt,autogen,CAMEL}, the agents and the workflows are designed manually according to one kind of task $t=P(q)$, as formulated $\mathcal{S}_{\text{task}} = \mathcal{H}(t)$, where $\mathcal{H}$ denotes human experts, and $\mathcal{S}_{\text{task}}$ denotes the \textbf{task-level multi-agent system} which is fixed for all queries in one task. This kind of manually designed multi-agent system leads to extensive human costs. Besides, such a one-size-fits-all system fails to allocate inference resources dynamically for different user queries within the task.

As illustrated in Figure \ref{fig:diff} (b), search-based automatic multi-agent systems \citep{AFLOW, ADAS, chen2023autoagents} are proposed to reduce human effort. Specifically, these approaches first prompt an LLM, such as GPT-4o, to act as a meta-agent, generating multiple candidate multi-agent system designs. Subsequently, complex search algorithms are employed on the carefully designed search set to identify the optimal system for completing the task. We denote this kind of meta-agent as \textbf{task-level meta-agent} $\mathcal{A}_{\text{meta\_task}}$ since they can merely generate one \textbf{task-level multi-agent system} $\mathcal{S}_{\text{task}}$ to solve one kind of task $t$. Although effective, they are still one-size-fits-all systems. Besides, the search algorithm is time-consuming and relies on the search set, which is absent in one user query.



\section{Meta-Agent \ourmethod}
\begin{figure*}[t]
\centering
\includegraphics[width=1.0\linewidth]{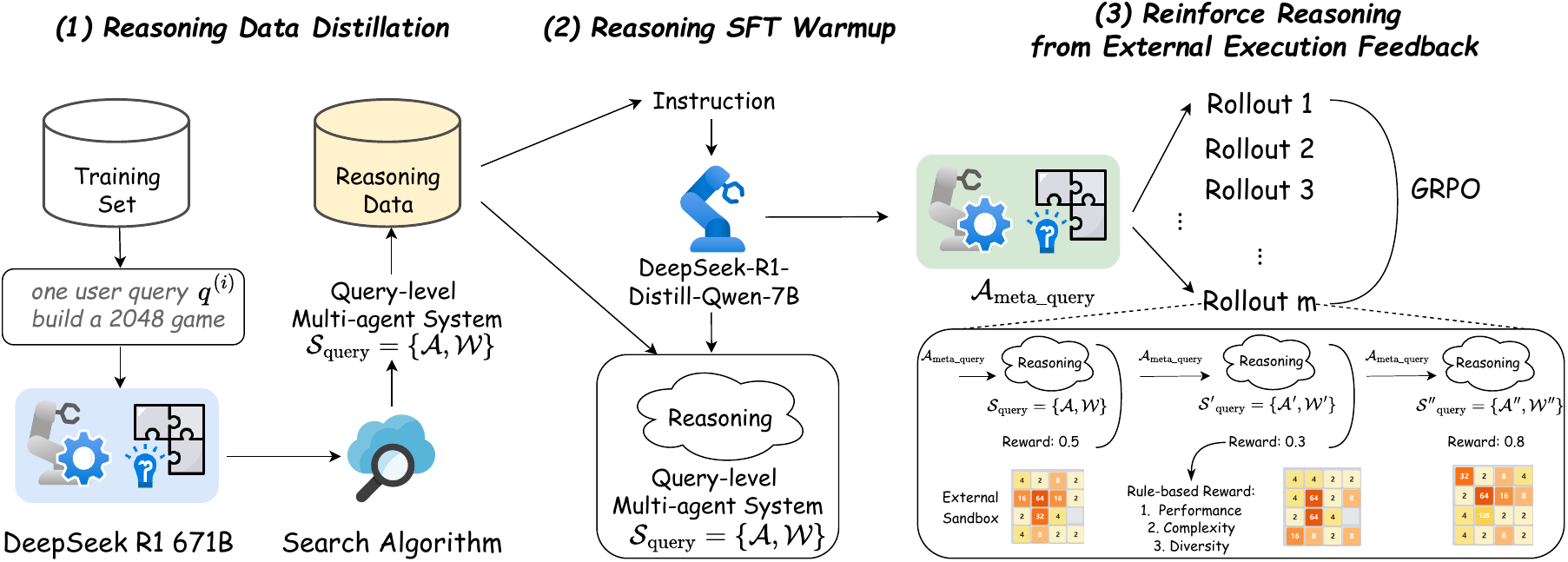}
   \caption{\textbf{Training Pipeline of \ourmethod.} It consists of (1) Reasoning Data Distillation, (2) Reasoning SFT Warmup, (3) Reinforce Reasoning from external execution feedback.} 
    \label{fig:ours_overview}
\end{figure*}

To solve these problems, we develop a reasoning-based automatic multi-agent system shown in Figure \ref{fig:diff} (c). Concretely, by guiding the model to reason from external execution feedback, we train a \textbf{query-level meta-agent} denoted as $\mathcal{A}_{\text{meta\_query}}$, which can automatically propose a novel \textbf{query-level multi-agent system} for each user query $q$, as formulated $\mathcal{S_\text{query}} = \mathcal{A}_{\text{meta\_query}}(q)$. Then, $\mathcal{S_\text{query}}$ accomplishes the specific user query and obtain the result $a$, i.e., $a = \mathcal{S_\text{query}}(q)$. Subsequently, the evaluator $\mathcal{E}$ evaluates the performance of the proposed multi-agents system $\mathcal{S_\text{query}}$ by comparing the result $a$ with ground truth $a_{\text{gt}}$ as formulated $\mathcal{E}(a,a_{\text{gt}})$. Our proposed method is more practical in real-world scenarios, as it can design an optimal multi-agent system for each specific query. The training pipeline of our method is demonstrated in Figure \ref{fig:ours_overview}.



\subsection{Learn to Reason}

\textbf{Reasoning Data Syntheses.} To enable our model to learn how to reason workflows based on external execution feedback, we first generate multi-round reasoning data using R1-671B\citep{deepseek_r1}. For a given input query $q$, the R1 model generates $l$ rounds of reasoning $\mathcal{R}_{\text{query}}$ and a multi-agent system $\mathcal{S}_{\text{query}}$ with external execution feedback at each round as follows:
\begin{equation} 
\{\{\mathcal{R}_{\text{query}}, \mathcal{S}_{\text{query}}\}^{(1)},...,\{\mathcal{R}_{\text{query}}, \mathcal{S}_{\text{query}}\}^{(l)}\} = \mathcal{A_\text{meta\_query}}(q).
\label{k_round}
\end{equation}
Then, we concatenate the model's $l$ rounds of reasoning and multi-agent system to form a reasoning process $\mathcal{\hat{R}}$ and final multi-agent system $\mathcal{\hat{S}}$, and pair it with the instruction $\mathcal{I}$ and query $q$ to construct our training sample $\{\mathcal{I}, q,\mathcal{\hat{R}},\mathcal{\hat{S}}\}$. Based on this, we construct a warmup SFT dataset $\mathcal{D}$.

\textbf{Reasoning SFT Warmup.} After creating the reasoning training dataset $\mathcal{D}$, we proceed to perform reasoning SFT for warmup. We input the instruction $\mathcal{I}$ and query $q$, then guide DeepSeek-R1-Distill-Qwen-7B to output reasoning process $\mathcal{\hat{R}}$ and final multi-agent system $\mathcal{\hat{S}}$. It can be formulated as follows:
\begin{equation} 
\mathcal{L}_{\text{SFT}} = -\mathbb{E}_{(\mathcal{I},q,\mathcal{\hat{R}},\mathcal{\hat{S}})\sim\mathcal{D}} \log P_{\theta}(\mathcal{\hat{R}},\mathcal{\hat{S}} \mid \mathcal{I},q),
\label{sft_loss}
\end{equation}
where $\theta$ denotes the model parameters. Through SFT, we unlock the model's reasoning ability regarding workflow generation.



\subsection{Reinforce Reasoning from External Execution Feedback}

After the SFT stage, we use reinforcement learning phase to further enhance the model's reasoning capabilities through workflows built on external execution feedback. We aims to leverage feedback to improve performance on complex tasks in multi-round multi-agent systems. Following the DeepSeek-R1-Zero~\citep{deepseek_r1}, we adopt standard GRPO (Grouped Relative Policy Optimization) as our training method. GRPO works by sampling multiple outputs for each query and computing relative advantages based on the rewards these outputs receive. The policy is then updated by maximizing an objective function incorporating these relative advantages. The GRPO objective function can be expressed as follows:
\begin{align}
  \mathcal{L}_{\mathrm{GRPO}}(\theta) &= \mathbb{E}_{q \sim t, \{o_i\}_{i=1}^G \sim \pi_{\theta_{\text{old}}}(O|q)} \notag \\
  &\quad \frac{1}{G} \sum_{i=1}^{G}  \left\{ 
  \min \left[ r_{\mathrm{ratio}}, \operatorname{clip}(r_{\mathrm{ratio}}, 1-\varepsilon, 1+\varepsilon)\right] 
  \cdot \hat{E}_{i,m} - \beta \mathbb{D}_{\mathrm{KL}}(\pi_{\theta} \| \pi_{\mathrm{ref}}) 
  \right\}\textrm{,}
\end{align}
where $G$ represents the number of sampled trajectories in each group, $o_i$ denotes the $i$-th trajectory, $|o_i|$ is the length of trajectory $o_i$, $r_{\text{ratio}} = \frac{\pi_{\theta}(o_{i,m} \mid q, o_{i,<m})}{\pi_{\theta_{\text{old}}}(o_{i,m} \mid q, o_{i,<m})}$ represents the probability ratio between new and old policies, $\hat{E}_{i,m}$ is the estimated advantage for the $m$-th token in the $i$-th trajectory, $\varepsilon$ is a small positive clipping parameter that prevents excessively large update steps, and the KL regularization term constrains policy drift to maintain stability during training.



Since we can obtain a reward from the external environment in every round, we use process reward supervision. The reward for each reasoning round is normalized, and the advantage function is calculated as the sum of subsequent normalized rewards.
\begin{equation}
\hat{E}_{i,m} = \sum_{j \geq T} \tilde{r}_{j}^{(i)}.
\end{equation}
Here, the normalized reward for the $j$-th round of the 
$i$-th candidate output is defined as $\tilde{r}_{j}^{(i)} = k^j \cdot \frac{r_{j}^{(i)} - \operatorname{mean}(\mathbf{R})}{\operatorname{std}(\mathbf{R})}$, 
where \(k\) is a scaling factor and $T$ is a threshold used to exclude the first $T$ items from the calculation. The set \(\mathbf{R}\) represents the list of scores for each round across all candidates, and $r_{j}^{(i)}$ is the score of candidate \(i\) in round \(j\). Each score $r_{j}^{(i)}$ is calculated by the performance of the proposed solution (i.e. pass rate), algorithm complexity (i.e. complexity score of abstract syntax tree) and diversity (i.e. distinctness ratio followed by \citet{overthinking}) of workflow.

\subsection{Generate Multi-Agent Systems with \ourmethod}

Constructing a multi-agent system is essentially an optimization problem with the goal of designing an optimized system, \(\mathcal{S}^*_{\text{query}}\), that responds to user queries. When a user submits a query \(q\), the system produces an answer $a = \mathcal{S}_{\text{query}}(q)$. The performance of the system in each round is evaluated using an external feedback function \(\mathcal{E}(a, a_{\textrm{gt}})\), where \(a_{\textrm{gt}}\) represents the ground truth answer. In this framework, a meta-agent, \(\mathcal{A}_{\text{meta\_query}}\), is responsible for optimizing the workflow with external execution feedback \(\mathcal{E}(a, a_{\textrm{gt}})\) in every round. The pass rate of the proposed solution for the given query, as the key performance indicator of the system, serves as the external execution feedback.
The optimization space is defined by all possible configurations of nodes and edges. Here, nodes represent various parameters (such as language models, prompts, temperature, and output formats), and edges capture the interactions or data flows between these nodes.

By representing both nodes and edges in code and employing predefined operators (such as Ensemble, Review, and Revise) along with a custom operator to combine these elements, \ourmethod utilizes an \(l\)-round optimization process same as Aflow~\citep{AFLOW} to arrive at the final multi-agent system:
\begin{equation}
    \mathcal{S}^*_{\text{query}} = \arg\max_{\mathcal{S}_{\text{query}} \in \mathcal{\textbf{S}}} \mathcal{E}(\mathcal{S}_{\text{query}}(q),a_{\textrm{gt}})\textrm{,}
\end{equation}
where \(\mathcal{S}^*_{\text{query}}\) is the optimal multi-agent system refined through optimization. By optimizing a multi-agent system through iterative external execution feedback, \ourmethod can construct a highly adaptive system that maximizes the accuracy and performance of solution for a query.





\begin{table}[t]
\centering
\caption{\textbf{Performance Evaluation.} Accuracy comparison across three code benchmarks for three categories of baselines - individual models, manual workflows, and automated workflow methods - alongside our \ourmethod-14B. For manual methods, model names in parentheses indicate the worker model used.}
\vspace{0.1cm}
\resizebox{0.96\textwidth}{!}{%
\begin{tabular}{cccccc}

\toprule
\textbf{Type}           & \textbf{Method}         & \textbf{BigCodeBench} & \textbf{HumanEval}   & \textbf{MBPP}     & \textbf{Overall}      \\
\midrule
\multirow{2}{*}{Vanilla} & o1-mini                 & 57.67                 & 95.42                & 74.19   &    71.37         \\
                         & GPT-4o-mini             & 56.33                 & 88.55                & 71.73    &   68.60         \\
\midrule
\multirow{6}{*}{Manual}  & Self-Refine (4o-mini) \citep{madaan2023self}     & 54.78                 & 89.83                & 69.64     &   67.29        \\
                         & LLM-Debate (4o-mini)  \citep{llm-debate}    & 56.88                 & 91.64                & 70.28     &  68.69         \\
                        & LLM-Blender (4o-mini) \citep{Llm-blender} & 57.46		   & 89.44             & 76.39               & 71.25              \\
                         & Self-Refine (o1-mini) \citep{madaan2023self}   & 56.68                 & 94.74                & 73.64    &   70.63         \\
                         & LLM-Debate (o1-mini) \citep{llm-debate}     & 57.25                 & 95.83                & 74.28    &  71.33          \\
                        & LLM-Blender (o1-mini) \citep{Llm-blender}  & 59.51              & 96.37              & 78.65      & 74.22        \\
                         
\midrule
\multirow{4}{*}{Auto}    & AutoAgents \citep{chen2023autoagents}                   & 56.65                    & 88.91                  & 72.03     &   68.92           \\
 & ADAS \citep{ADAS}                  & 53.87		                & 84.26                 & 68.47      &   65.48          \\
                         & Aflow \citep{AFLOW}                   & 59.83                 & 94.15                & 82.40     &  75.63         \\
                         & MaAS \citep{MaAS}                   & 60.33                 & 95.42                & 84.16     &  76.81         \\
\midrule
\multirow{1}{*}{Ours}  
                   \cellcolor{gray!25}      & \cellcolor{gray!25}\ourmethod-14B & \cellcolor{gray!25}\textbf{63.53} & \cellcolor{gray!25}\textbf{97.26} & \cellcolor{gray!25}\textbf{92.15} & \cellcolor{gray!25}\textbf{81.89}\\
\bottomrule
\end{tabular}}

\label{main}
\end{table}

\section{Experiments}
\begin{table}[t]
\centering
\caption{\textbf{Ablation Study on Model Sizes and Training Stages.} Accuracy(\%) comparison across three code benchmarks for models of different sizes (7B/14B) at both the Supervised Fine-Tuning (SFT) stage and the combined SFT with Reinforcement Learning (SFT+RL) stage.}
\label{Model_Sizes}
\vspace{0.1cm}
\resizebox{0.7\textwidth}{!}{%
\begin{tabular}{cccccc}

\toprule
 \textbf{Stage}&  \textbf{Size}       & \textbf{BigCodeBench} & \textbf{HumanEval}   & \textbf{MBPP}     & \textbf{Overall}      \\
\midrule
   SFT&  7B  & 61.79                 & 96.38                & 87.22    &   78.89         \\
                          SFT+RL& 7B  & 62.78                 & 96.95                & 89.86   &  80.53           \\
                         SFT&  14B  & 62.83                 & 97.18              & 91.91  &  81.50            \\
 SFT+RL& 14B & 63.53 & 97.26 & 92.15 & 81.89\\
\bottomrule
\end{tabular}}
\end{table}

\begin{figure*}[t]
\centering
    \centering
    \begin{subfigure}[b]{0.638\textwidth}
        \centering
        \includegraphics[width=\textwidth]{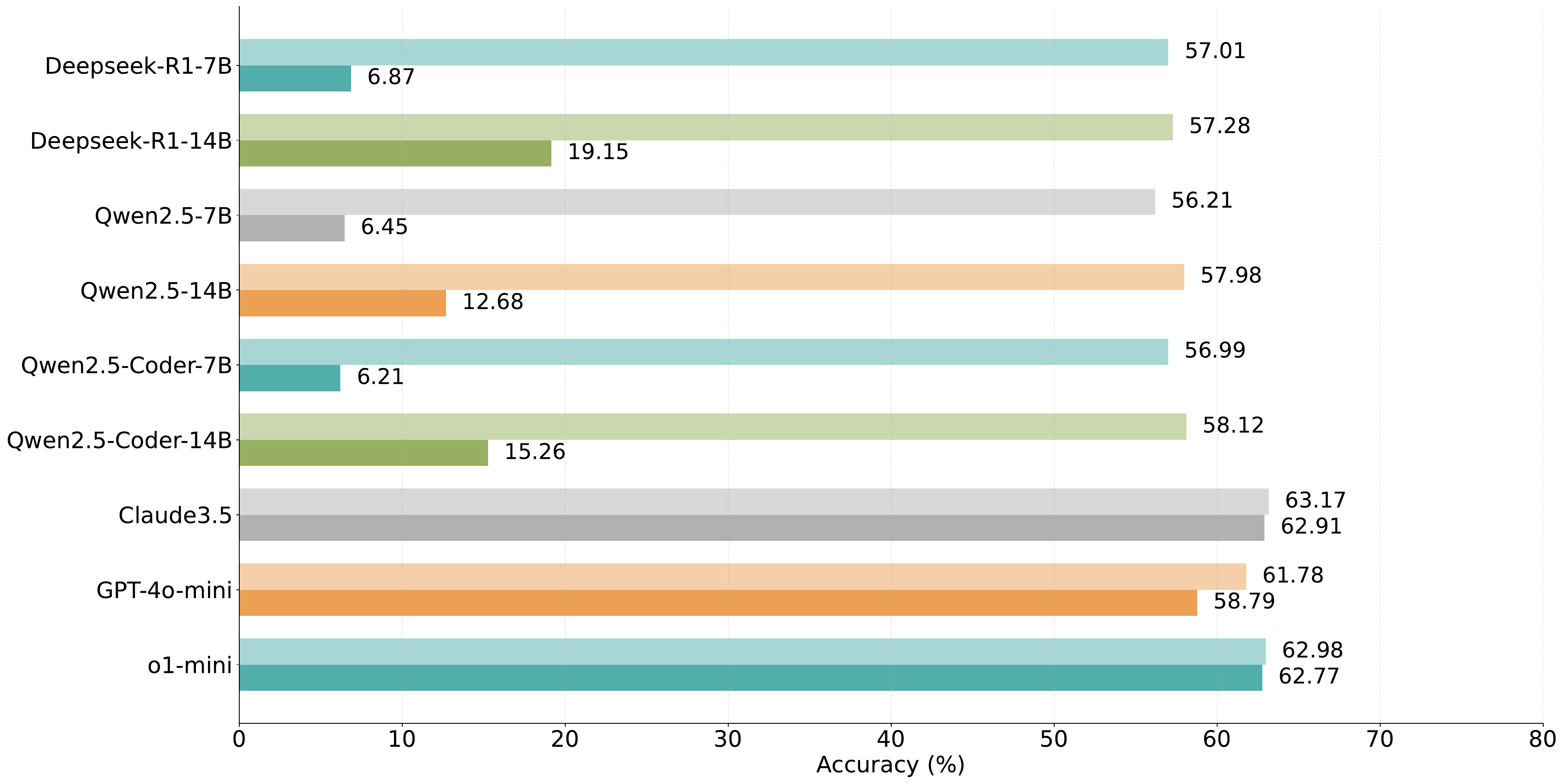}
        \caption{Ablation of Meta-agent}
        \label{hit:subfig1}
    \end{subfigure}
    \hfill
    \begin{subfigure}[b]{0.354\textwidth}
        \centering
        \includegraphics[width=\textwidth]{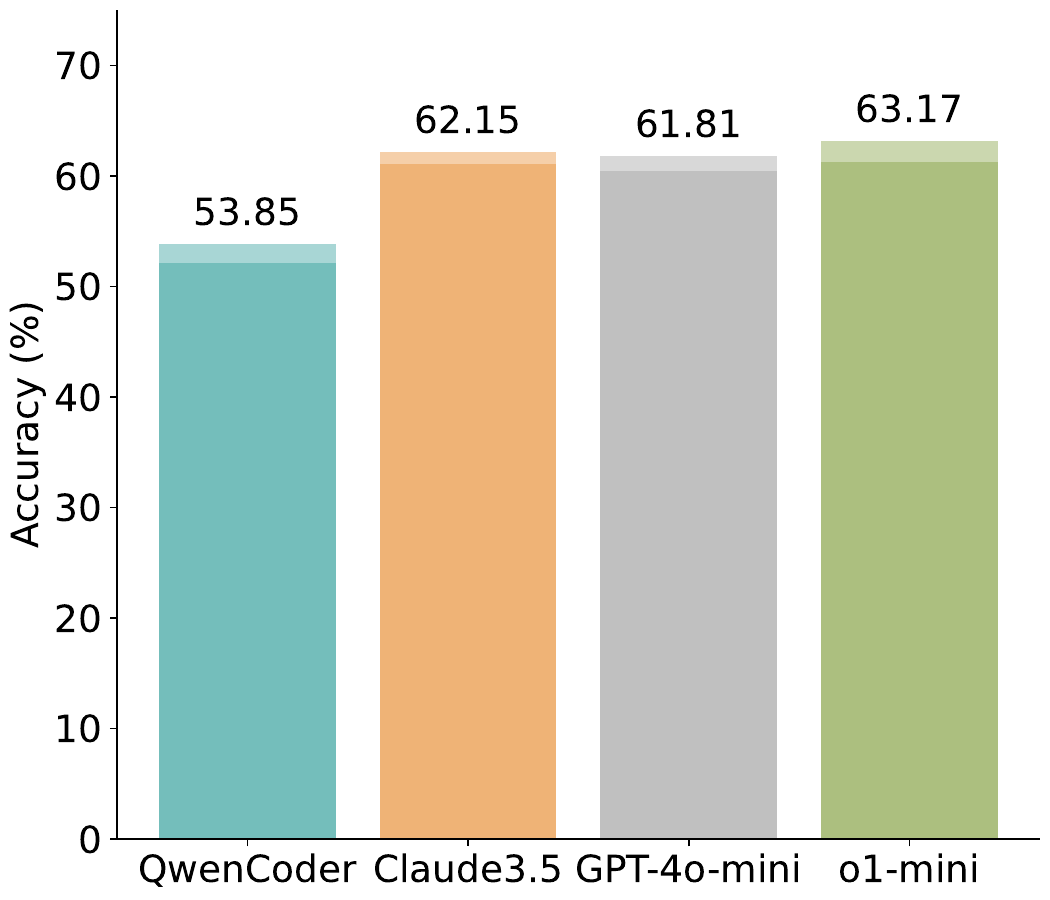}
        \caption{Ablation of Workers}
        \label{hit:subfig2}
    \end{subfigure}
    \caption{\textbf{Ablation of Meta-agent and Workers.} (a) Accuracy of different meta-agents with o1-mini as workers. (B) Accuracy of the generated workflow with different workers.}
    \label{fig:whole1}
\end{figure*}

\begin{figure}[t]
    \centering
    \vspace{-0.2cm}
    \begin{subfigure}{0.95\textwidth}
        \centering
        \begin{lstlisting}[language=Python,  basicstyle=\scriptsize\ttfamily]
async def __call__(self, problem: str, test_cases: list):
    max_retries = 3
    solutions = []

    for _ in range(max_retries):
        for attempt in range(3):  
        # Try generating code up to 3 times
            solution = await self.custom_code_generate(
                problem=problem, instruction=prompt_custom.GENERATE_CODE_PROMPT
                )
            test_result = await self.test(
                problem=problem, solution=solution['response'], test_cases=test_cases
            )
            
            if test_result['result']:
                solutions.append(solution['response'])
                break  # Break the inner loop if a correct solution is found
            elif attempt < 2:  
            # If it's not the last attempt, try to improve the solution
                error_analysis = await self.custom(
                input=f"Problem: {problem}\nFailed solution: {solution['response']}\nError: {test_result['solution']}", instruction=prompt_custom.ERROR_ANALYSIS_PROMPT
                )
                prompt_custom.GENERATE_CODE_PROMPT = f"{prompt_custom.GENERATE_CODE_PROMPT}\n{error_analysis['response']}"
        
        if not solutions:  # If no correct solution was found after 3 attempts
            solutions.append(solution['response'])  # Add the last generated solution

    best_solution = await self.sc_ensemble(solutions=solutions, problem=problem)
    return best_solution['response'], self.llm.cost_manager.total_cost
        \end{lstlisting}
        \caption{Workflow generated by \ourmethod-14B for the BigCodeBench task of \emph{generating traffic data for various vehicle types over a specified number of hours, saving the data to a CSV file with columns, and plotting it in a line chart.}}
         \vspace{2pt}
        \label{fig:python-code}
    \end{subfigure}
    \hfill
    \begin{subfigure}{0.95\textwidth}
        \centering
        \begin{lstlisting}[language=Python,  basicstyle=\scriptsize\ttfamily]
    async def __call__(self, problem: str, test_cases: list):
        solution = await self.custom_code_generate(problem=problem, entry_point=entry_point, instruction=prompt_custom.CODE_GENERATE_PROMPT)
        
        # Add review to check the solution
        review = await self.custom(
        input=f"Problem: {problem}\nGenerated solution: {solution['response']}", instruction=prompt_custom.REVIEW_PROMPT
        )
        
        # Improve the solution based on the review information
        improved_solution = await self.custom_code_generate(
        problem=problem, instruction=f"{prompt_custom.CODE_IMPROVE_PROMPT}\n\nConsider this review: {review['response']}"
        )

        return improved_solution['response'], self.llm.cost_manager.total_cost
        \end{lstlisting}
        \caption{Workflow generated by \ourmethod-14B for the HumanEval task of \emph{spliting the string into words and returning an array of the words.}}
        \label{fig:cpp-code}
    \end{subfigure}
    
    \caption{Cases of Workflows generated by \ourmethod-14B for the tasks of BigCodeBench and HumanEval.}
    \label{fig:code-examples}
\end{figure}

\textbf{Datasets.}
Given our focus on creating workflows tailored to individual user queries rather than the general task, we restrict our scope to \emph{code generation tasks}, as they can provide test cases as external execution feedback for the workflow construction process. Among various benchmarks, we consider three representative datasets: BigCodeBench\cite{bigcodebench}, which emphasizes engineering-oriented tasks, and two algorithmically focused benchmarks, HumanEval\citep{chen2021evaluating} and MBPP~\citep{austin2021program}. This selection enables us to comprehensively evaluate workflow discovery across a diverse spectrum of code generation challenges.\looseness=-1



\textbf{Baselines.}
In line with prior work~\citep{ADAS, AFLOW}, we evaluate \textsc{\ourmethod} against three categories of baselines: (1) single-model direct invocation, where a single LLM is prompted to solve the problem without additional structure; (2) manually designed workflows, including Self-Refine~\citep{madaan2023self}, LLM-Debate~\citep{llm-debate}, and LLM-Blender~\citep{Llm-blender}, which incorporate human-crafted reasoning strategies; and (3) automated workflow optimization methods, such as Aflow~\citep{AFLOW}, ADAS~\citep{ADAS}, and MaAS~\citep{MaAS}, which construct workflows through search or optimization over possible reasoning structures. This comparison enables a comprehensive assessment of \textsc{\ourmethod}’s effectiveness relative to both static and adaptive baselines.


\textbf{Implementation Details.}
For the manually designed workflow baselines, we employed both o1-mini and GPT-4o-mini as worker models for each method. For the automated workflow optimization baselines, we adopted the original configurations as described in MaAS~\citep{MaAS}. In our proposed method, \ourmethod, we trained two variants of DeepSeek-R1-Distill-Qwen with 7B and 14B parameters, respectively, and used o1-mini as the worker model. We fixed the number of workflow iterations to 10. To assess performance, we used the standard pass@1 metric for code accuracy, consistent with prior work~\citep{chen2021evaluating}. More details can be found in Appendix~\ref{details}.


\subsection{Experiment Results}

\textbf{Performance Comparison.}
Table~\ref{main} presents the performance comparison between our proposed method and the baselines. \ourmethod-14B consistently outperforms all competing approaches across the three benchmark datasets. Notably, it achieves an overall improvement of 5 percentage points over the strongest baseline, MaAS, and exceeds the performance of its underlying worker model, o1-mini, by a substantial margin of 10\%. These results highlight the effectiveness of our workflow-based reasoning framework in enhancing code generation accuracy.


\textbf{Ablation on Model Size and Training Stages.}
To investigate the impact of model size and training stages on performance, we conducted an ablation study comparing different configurations of \ourmethod in Table~\ref{Model_Sizes}. We observed that the 14B variant consistently outperformed the 7B counterpart across all benchmarks, indicating a positive correlation between model scale and reasoning effectiveness. Furthermore, within each model size, versions trained with both SFT and RL exhibited notable improvements over those trained with SFT alone, demonstrating the complementary benefits of incorporating RL in enhancing workflow-guided code generation.


\textbf{Ablation of Meta-agents and Workers.}
To analyze the impact of meta-agent and worker selection, we conducted an ablation study on the BigCodeBench dataset. Figure~\ref{main} presents performance comparisons under various meta-agent and worker configurations. As shown in Figure~\ref{fig:whole1} (a), open-source models exhibited poor performance when paired with o1-mini as the worker and no initial workflow, frequently generating error-prone workflows. This highlights a limitation of current open-source models, which struggle to produce reliable workflows solely based on instruction prompts and are heavily reliant on predefined, manually crafted workflows. In contrast, API-based models demonstrated stronger performance, likely attributable to their superior instruction-following capabilities. Figure~\ref{fig:whole1} (b) further examines worker effectiveness when using a high-performing meta-agent (Claude 3.5). We compared the open-source model Qwen2.5-Coder-32B and three API-based models, including Claude 3.5, GPT-4o-mini, and o1-mini. Among these, o1-mini achieved the best overall performance, suggesting its suitability as a worker model in the \ourmethod framework.

\begin{table}[t]
\centering
\caption{\textbf{Generalization Evaluation.} Accuracy of our trained meta-agent \ourmethod-7B/14B when paired with alternative workers including Qwen2.5 Coder, Claude, and GPT4o-mini.}
\vspace{0.1cm}
\resizebox{\textwidth}{!}{%
\begin{tabular}{ccccccc}
\toprule
\multirow{2}{*}{\textbf{Worker/Meta-agent}} & \multicolumn{3}{c}{\textbf{\ourmethod-7B (SFT+RL)}}       & \multicolumn{3}{c}{\textbf{\ourmethod-14B (SFT+RL)}}      \\
\cmidrule(lr){2-4} \cmidrule(lr){5-7}
                                           & \textbf{BigCodeBench} & \textbf{HumanEval} & \textbf{MBPP} & \textbf{BigCodeBench} & \textbf{HumanEval} & \textbf{MBPP} \\
                                           \midrule
\textbf{o1-mini}                           & 62.77                 & 96.95              & 89.86         & 63.53                 & 97.26              & 92.15         \\
\midrule
\textbf{Qwen2.5Coder-32b}                  & 50.17                 & 92.89              & 80.40         & 52.67                 & 93.69              & 78.90         \\
\textbf{Claude3.5}                         & 60.67                 & 96.07              & 87.63         & 61.12                 & 96.52              & 89.82         \\
\textbf{GPT-4o-mini}                       & 59.18                 & 94.24              & 82.19         & 59.75                 & 94.52              & 82.27         \\
\bottomrule
\end{tabular}%
}

\label{tab:gen}
\end{table}


\textbf{Generalization Capability.}
To evaluate whether \textsc{\ourmethod}, trained with o1-mini as the worker, can generalize its planning capabilities to alternative workers, we conducted experiments by substituting the worker with Qwen2.5-Coder, Claude, and GPT-4o-mini, while keeping the meta-agent fixed as either \ourmethod-7B or \ourmethod-14B. As shown in Table~\ref{tab:gen}, \ourmethod exhibits a notable degree of transferability, maintaining consistent performance across different worker models on the same tasks. These results suggest that the planner is not tightly bound to a specific worker and can adapt its strategies effectively across diverse execution agents.

\textbf{Case Study.}
Figure~\ref{fig:code-examples} illustrates two example workflows generated by \ourmethod-14B for representative tasks from BigCodeBench and HumanEval, respectively. The workflow corresponding to the BigCodeBench task exhibits greater complexity, reflecting the more challenging and engineering-oriented nature of the task. In contrast, the HumanEval workflow is substantially more concise, aligning with the relative simplicity and algorithmic focus of the task. These examples demonstrate \ourmethod’s ability to adapt the structure and granularity of workflows based on task complexity. More cases can be found in Appendix~\ref{case_more}.

\vspace{-0.05cm}
\section{Conclusion}
\vspace{-0.05cm}

In this paper, we present \ourmethod, a query-level meta-agent designed to automate the creation of personalized multi-agent systems for individual user queries. Unlike previous task-level approaches that create fixed, one-size-fits-all systems, \ourmethod dynamically generates tailored workflows for each specific query through reasoning-based optimization. The system leverages external execution feedback and reinforcement learning with multi-purpose rewards focusing on performance, complexity, and efficiency to generate optimized workflows without relying on complex search algorithms or carefully designed search sets. Experimental results demonstrate that \ourmethod-14B outperforms both manually designed workflows and existing automated methods across multiple code generation benchmarks, notably improving o1-mini's performance by overall 10.52\% on three benchmarks, thus proving the effectiveness and adaptability of reasoning-driven workflow generation. Besides, the observation when pairing \ourmethod with different worker models further confirms its generalization capabilities. Our approach reduces human resource costs while enhancing scalability by enabling more adaptive and efficient multi-agent systems that dynamically optimize their structure based on specific user queries rather than relying on fixed workflows for entire task categories.

\bibliography{ms}

\begin{thebibliography}{65}
\providecommand{\natexlab}[1]{#1}
\providecommand{\url}[1]{\texttt{#1}}
\expandafter\ifx\csname urlstyle\endcsname\relax
  \providecommand{\doi}[1]{doi: #1}\else
  \providecommand{\doi}{doi: \begingroup \urlstyle{rm}\Url}\fi

\bibitem[Achiam et~al.(2023)Achiam, Adler, Agarwal, Ahmad, Akkaya, Aleman, Almeida, Altenschmidt, Altman, Anadkat, et~al.]{gpt_4}
Josh Achiam, Steven Adler, Sandhini Agarwal, Lama Ahmad, Ilge Akkaya, Florencia~Leoni Aleman, Diogo Almeida, Janko Altenschmidt, Sam Altman, Shyamal Anadkat, et~al.
\newblock Gpt-4 technical report.
\newblock \emph{arXiv preprint arXiv:2303.08774}, 2023.

\bibitem[Anthropic(2024)]{computer_use}
Anthropic.
\newblock Introducing computer use, a new claude 3.5 sonnet, and claude 3.5 haiku.
\newblock \emph{https://www.anthropic.com/news/3-5-models-and-computer-use}, 2024.

\bibitem[ARC-AGI(2024)]{ARC_AGI}
ARC-AGI.
\newblock Abstraction and reasoning corpus for artificial general intelligence.
\newblock \emph{https://github.com/fchollet/ARC-AGI/}, 2024.

\bibitem[Austin et~al.(2021)Austin, Odena, Nye, Bosma, Michalewski, Dohan, Jiang, Cai, Terry, Le, et~al.]{austin2021program}
Jacob Austin, Augustus Odena, Maxwell Nye, Maarten Bosma, Henryk Michalewski, David Dohan, Ellen Jiang, Carrie Cai, Michael Terry, Quoc Le, et~al.
\newblock Program synthesis with large language models.
\newblock \emph{arXiv preprint arXiv:2108.07732}, 2021.

\bibitem[Chen et~al.(2023{\natexlab{a}})Chen, Dong, Shu, Zhang, Sesay, Karlsson, Fu, and Shi]{chen2023autoagents}
Guangyao Chen, Siwei Dong, Yu~Shu, Ge~Zhang, Jaward Sesay, B{\"o}rje~F Karlsson, Jie Fu, and Yemin Shi.
\newblock Autoagents: A framework for automatic agent generation.
\newblock \emph{arXiv preprint arXiv:2309.17288}, 2023{\natexlab{a}}.

\bibitem[Chen et~al.(2021)Chen, Tworek, Jun, Yuan, Pinto, Kaplan, Edwards, Burda, Joseph, Brockman, et~al.]{chen2021evaluating}
Mark Chen, Jerry Tworek, Heewoo Jun, Qiming Yuan, Henrique Ponde De~Oliveira Pinto, Jared Kaplan, Harri Edwards, Yuri Burda, Nicholas Joseph, Greg Brockman, et~al.
\newblock Evaluating large language models trained on code.
\newblock \emph{arXiv preprint arXiv:2107.03374}, 2021.

\bibitem[Chen et~al.(2023{\natexlab{b}})Chen, Su, Zuo, Yang, Yuan, Qian, Chan, Qin, Lu, Xie, et~al.]{agentverse}
Weize Chen, Yusheng Su, Jingwei Zuo, Cheng Yang, Chenfei Yuan, Chen Qian, Chi-Min Chan, Yujia Qin, Yaxi Lu, Ruobing Xie, et~al.
\newblock Agentverse: Facilitating multi-agent collaboration and exploring emergent behaviors in agents.
\newblock \emph{arXiv preprint arXiv:2308.10848}, 2\penalty0 (4):\penalty0 6, 2023{\natexlab{b}}.

\bibitem[Chen et~al.(2024)Chen, Xu, Liang, He, Pang, Yu, Song, Liu, Zhou, Zhang, et~al.]{overthinking}
Xingyu Chen, Jiahao Xu, Tian Liang, Zhiwei He, Jianhui Pang, Dian Yu, Linfeng Song, Qiuzhi Liu, Mengfei Zhou, Zhuosheng Zhang, et~al.
\newblock Do not think that much for 2+ 3=? on the overthinking of o1-like llms.
\newblock \emph{arXiv preprint arXiv:2412.21187}, 2024.

\bibitem[CognitionAI(2024)]{devin}
CognitionAI.
\newblock Introducing devin, the first ai software engineer.
\newblock \emph{https://www.cognition.ai/blog/introducing-devin/}, 2024.

\bibitem[Du et~al.(2023)Du, Li, Torralba, Tenenbaum, and Mordatch]{llm-debate}
Yilun Du, Shuang Li, Antonio Torralba, Joshua~B Tenenbaum, and Igor Mordatch.
\newblock Improving factuality and reasoning in language models through multiagent debate, 2023.
\newblock \emph{URL https://arxiv. org/abs/2305.14325}, 3, 2023.

\bibitem[Feng et~al.(2025)Feng, Wang, Goyal, Wang, Shi, Xia, Palangi, Zettlemoyer, Tsvetkov, Lee, et~al.]{feng2025heterogeneous}
Shangbin Feng, Zifeng Wang, Palash Goyal, Yike Wang, Weijia Shi, Huang Xia, Hamid Palangi, Luke Zettlemoyer, Yulia Tsvetkov, Chen-Yu Lee, et~al.
\newblock Heterogeneous swarms: Jointly optimizing model roles and weights for multi-llm systems.
\newblock \emph{arXiv preprint arXiv:2502.04510}, 2025.

\bibitem[Fernando et~al.(2023)Fernando, Banarse, Michalewski, Osindero, and Rockt{\"a}schel]{fernando2023promptbreeder}
Chrisantha Fernando, Dylan Banarse, Henryk Michalewski, Simon Osindero, and Tim Rockt{\"a}schel.
\newblock Promptbreeder: Self-referential self-improvement via prompt evolution.
\newblock \emph{arXiv preprint arXiv:2309.16797}, 2023.

\bibitem[Gao et~al.(2025)Gao, Qu, Tang, Bi, Liu, Chen, Liang, Su, and Huang]{gao2025exploring}
Hongcheng Gao, Jiashu Qu, Jingyi Tang, Baolong Bi, Yue Liu, Hongyu Chen, Li~Liang, Li~Su, and Qingming Huang.
\newblock Exploring hallucination of large multimodal models in video understanding: Benchmark, analysis and mitigation.
\newblock \emph{arXiv preprint arXiv:2503.19622}, 2025.

\bibitem[Goyal et~al.(2023)Goyal, Ji, Rawat, Menon, Kumar, and Nagarajan]{pause_token}
Sachin Goyal, Ziwei Ji, Ankit~Singh Rawat, Aditya~Krishna Menon, Sanjiv Kumar, and Vaishnavh Nagarajan.
\newblock Think before you speak: Training language models with pause tokens.
\newblock \emph{arXiv preprint arXiv:2310.02226}, 2023.

\bibitem[Guo et~al.(2024)Guo, Chen, Wang, Chang, Pei, Chawla, Wiest, and Zhang]{survey}
Taicheng Guo, Xiuying Chen, Yaqi Wang, Ruidi Chang, Shichao Pei, Nitesh~V Chawla, Olaf Wiest, and Xiangliang Zhang.
\newblock Large language model based multi-agents: A survey of progress and challenges.
\newblock \emph{arXiv preprint arXiv:2402.01680}, 2024.

\bibitem[Hao et~al.(2024)Hao, Sukhbaatar, Su, Li, Hu, Weston, and Tian]{latent_space}
Shibo Hao, Sainbayar Sukhbaatar, DiJia Su, Xian Li, Zhiting Hu, Jason Weston, and Yuandong Tian.
\newblock Training large language models to reason in a continuous latent space.
\newblock \emph{arXiv preprint arXiv:2412.06769}, 2024.

\bibitem[Hong et~al.(2023)Hong, Zheng, Chen, Cheng, Wang, Zhang, Wang, Yau, Lin, Zhou, et~al.]{metagpt}
Sirui Hong, Xiawu Zheng, Jonathan Chen, Yuheng Cheng, Jinlin Wang, Ceyao Zhang, Zili Wang, Steven Ka~Shing Yau, Zijuan Lin, Liyang Zhou, et~al.
\newblock Metagpt: Meta programming for multi-agent collaborative framework.
\newblock \emph{arXiv preprint arXiv:2308.00352}, 3\penalty0 (4):\penalty0 6, 2023.

\bibitem[Hu et~al.(2024)Hu, Lu, and Clune]{ADAS}
Shengran Hu, Cong Lu, and Jeff Clune.
\newblock Automated design of agentic systems.
\newblock \emph{arXiv preprint arXiv:2408.08435}, 2024.

\bibitem[Huang et~al.(2024)Huang, Feng, Li, Xiang, Wang, Liu, and Qin]{yichong_nips}
Yichong Huang, Xiaocheng Feng, Baohang Li, Yang Xiang, Hui Wang, Ting Liu, and Bing Qin.
\newblock Ensemble learning for heterogeneous large language models with deep parallel collaboration.
\newblock In A.~Globerson, L.~Mackey, D.~Belgrave, A.~Fan, U.~Paquet, J.~Tomczak, and C.~Zhang, editors, \emph{Advances in Neural Information Processing Systems}, volume~37, pages 119838--119860. Curran Associates, Inc., 2024.

\bibitem[Jiang et~al.(2023)Jiang, Ren, and Lin]{Llm-blender}
Dongfu Jiang, Xiang Ren, and Bill~Yuchen Lin.
\newblock Llm-blender: Ensembling large language models with pairwise ranking and generative fusion.
\newblock \emph{arXiv preprint arXiv:2306.02561}, 2023.

\bibitem[Ke et~al.(2023)Ke, Wen, Feng, Liu, Lei, Cheng, Wang, Zeng, Dong, Wang, et~al.]{self_critique}
Pei Ke, Bosi Wen, Zhuoer Feng, Xiao Liu, Xuanyu Lei, Jiale Cheng, Shengyuan Wang, Aohan Zeng, Yuxiao Dong, Hongning Wang, et~al.
\newblock Critiquellm: Scaling llm-as-critic for effective and explainable evaluation of large language model generation.
\newblock \emph{arXiv preprint arXiv:2311.18702}, 2023.

\bibitem[Khattab et~al.(2024)Khattab, Singhvi, Maheshwari, Zhang, Santhanam, Haq, Sharma, Joshi, Moazam, Miller, et~al.]{Dspy}
Omar Khattab, Arnav Singhvi, Paridhi Maheshwari, Zhiyuan Zhang, Keshav Santhanam, Saiful Haq, Ashutosh Sharma, Thomas~T Joshi, Hanna Moazam, Heather Miller, et~al.
\newblock Dspy: Compiling declarative language model calls into state-of-the-art pipelines.
\newblock In \emph{The Twelfth International Conference on Learning Representations}, 2024.

\bibitem[Kim et~al.(2024)Kim, Pertsch, Karamcheti, Xiao, Balakrishna, Nair, Rafailov, Foster, Lam, Sanketi, et~al.]{openvla}
Moo~Jin Kim, Karl Pertsch, Siddharth Karamcheti, Ted Xiao, Ashwin Balakrishna, Suraj Nair, Rafael Rafailov, Ethan Foster, Grace Lam, Pannag Sanketi, et~al.
\newblock Openvla: An open-source vision-language-action model.
\newblock \emph{arXiv preprint arXiv:2406.09246}, 2024.

\bibitem[Kojima et~al.(2022)Kojima, Gu, Reid, Matsuo, and Iwasawa]{kojima2022large}
Takeshi Kojima, Shixiang~Shane Gu, Machel Reid, Yutaka Matsuo, and Yusuke Iwasawa.
\newblock Large language models are zero-shot reasoners.
\newblock \emph{Advances in neural information processing systems}, 35:\penalty0 22199--22213, 2022.

\bibitem[Kumar et~al.(2024)Kumar, Zhuang, Agarwal, Su, Co-Reyes, Singh, Baumli, Iqbal, Bishop, Roelofs, et~al.]{self_correct_1}
Aviral Kumar, Vincent Zhuang, Rishabh Agarwal, Yi~Su, John~D Co-Reyes, Avi Singh, Kate Baumli, Shariq Iqbal, Colton Bishop, Rebecca Roelofs, et~al.
\newblock Training language models to self-correct via reinforcement learning.
\newblock \emph{arXiv preprint arXiv:2409.12917}, 2024.

\bibitem[Li et~al.(2023)Li, Hammoud, Itani, Khizbullin, and Ghanem]{CAMEL}
Guohao Li, Hasan Hammoud, Hani Itani, Dmitrii Khizbullin, and Bernard Ghanem.
\newblock Camel: Communicative agents for" mind" exploration of large language model society.
\newblock \emph{Advances in Neural Information Processing Systems}, 36:\penalty0 51991--52008, 2023.

\bibitem[Li et~al.(2024)Li, Xu, Mei, Hua, Rama, Raheja, Wang, Zhu, and Zhang]{li2024autoflow}
Zelong Li, Shuyuan Xu, Kai Mei, Wenyue Hua, Balaji Rama, Om~Raheja, Hao Wang, He~Zhu, and Yongfeng Zhang.
\newblock Autoflow: Automated workflow generation for large language model agents.
\newblock \emph{arXiv preprint arXiv:2407.12821}, 2024.

\bibitem[Li et~al.(2025)Li, Zhang, Zhang, Zhang, Liu, Yao, Xu, Zheng, Wang, Chen, et~al.]{li2025system}
Zhong-Zhi Li, Duzhen Zhang, Ming-Liang Zhang, Jiaxin Zhang, Zengyan Liu, Yuxuan Yao, Haotian Xu, Junhao Zheng, Pei-Jie Wang, Xiuyi Chen, et~al.
\newblock From system 1 to system 2: A survey of reasoning large language models.
\newblock \emph{arXiv preprint arXiv:2502.17419}, 2025.

\bibitem[Liang et~al.(2023)Liang, He, Jiao, Wang, Wang, Wang, Yang, Shi, and Tu]{debate_1}
Tian Liang, Zhiwei He, Wenxiang Jiao, Xing Wang, Yan Wang, Rui Wang, Yujiu Yang, Shuming Shi, and Zhaopeng Tu.
\newblock Encouraging divergent thinking in large language models through multi-agent debate.
\newblock \emph{arXiv preprint arXiv:2305.19118}, 2023.

\bibitem[Liu et~al.(2024)Liu, Feng, Xue, Wang, Wu, Lu, Zhao, Deng, Zhang, Ruan, et~al.]{deepseek_v3}
Aixin Liu, Bei Feng, Bing Xue, Bingxuan Wang, Bochao Wu, Chengda Lu, Chenggang Zhao, Chengqi Deng, Chenyu Zhang, Chong Ruan, et~al.
\newblock Deepseek-v3 technical report.
\newblock \emph{arXiv preprint arXiv:2412.19437}, 2024.

\bibitem[Liu et~al.(2025{\natexlab{a}})Liu, Gao, Zhai, Xia, Wu, Xue, Chen, Kawaguchi, Zhang, and Hooi]{liu2025guardreasoner}
Yue Liu, Hongcheng Gao, Shengfang Zhai, Jun Xia, Tianyi Wu, Zhiwei Xue, Yulin Chen, Kenji Kawaguchi, Jiaheng Zhang, and Bryan Hooi.
\newblock Guardreasoner: Towards reasoning-based llm safeguards.
\newblock \emph{arXiv preprint arXiv:2501.18492}, 2025{\natexlab{a}}.

\bibitem[Liu et~al.(2025{\natexlab{b}})Liu, Wu, He, Gao, Chen, Bi, Zhang, Huang, and Hooi]{token_efficiency_survey}
Yue Liu, Jiaying Wu, Yufei He, Hongcheng Gao, Hongyu Chen, Baolong Bi, Jiaheng Zhang, Zhiqi Huang, and Bryan Hooi.
\newblock Efficient inference for large reasoning models: A survey.
\newblock \emph{arXiv preprint arXiv:2503.23077}, 2025{\natexlab{b}}.

\bibitem[Liu et~al.(2023)Liu, Zhang, Li, Liu, and Yang]{DyLAN}
Zijun Liu, Yanzhe Zhang, Peng Li, Yang Liu, and Diyi Yang.
\newblock Dynamic llm-agent network: An llm-agent collaboration framework with agent team optimization.
\newblock \emph{arXiv preprint arXiv:2310.02170}, 2023.

\bibitem[Madaan et~al.(2023)Madaan, Tandon, Gupta, Hallinan, Gao, Wiegreffe, Alon, Dziri, Prabhumoye, Yang, et~al.]{madaan2023self}
Aman Madaan, Niket Tandon, Prakhar Gupta, Skyler Hallinan, Luyu Gao, Sarah Wiegreffe, Uri Alon, Nouha Dziri, Shrimai Prabhumoye, Yiming Yang, et~al.
\newblock Self-refine: Iterative refinement with self-feedback.
\newblock \emph{Advances in Neural Information Processing Systems}, 36:\penalty0 46534--46594, 2023.

\bibitem[OpenAI(2022)]{chatgpt}
OpenAI.
\newblock Introducing chatgpt.
\newblock \emph{https://openai.com/index/chatgpt/}, 2022.

\bibitem[OpenAI(2024)]{o1}
OpenAI.
\newblock Learning to reason with llms.
\newblock \emph{https://openai.com/index/learning-to-reason-with-llms/}, 2024.

\bibitem[OpenAI(2025)]{deep_research}
OpenAI.
\newblock Introducing deep research.
\newblock \emph{https://openai.com/index/introducing-deep-research/}, 2025.

\bibitem[Park et~al.(2023)Park, O'Brien, Cai, Morris, Liang, and Bernstein]{park2023generative}
Joon~Sung Park, Joseph O'Brien, Carrie~Jun Cai, Meredith~Ringel Morris, Percy Liang, and Michael~S Bernstein.
\newblock Generative agents: Interactive simulacra of human behavior.
\newblock In \emph{Proceedings of the 36th annual acm symposium on user interface software and technology}, pages 1--22, 2023.

\bibitem[Rafailov et~al.(2023)Rafailov, Sharma, Mitchell, Manning, Ermon, and Finn]{dpo}
Rafael Rafailov, Archit Sharma, Eric Mitchell, Christopher~D Manning, Stefano Ermon, and Chelsea Finn.
\newblock Direct preference optimization: Your language model is secretly a reward model.
\newblock \emph{Advances in Neural Information Processing Systems}, 36:\penalty0 53728--53741, 2023.

\bibitem[Reid et~al.(2024)Reid, Savinov, Teplyashin, Lepikhin, Lillicrap, Alayrac, Soricut, Lazaridou, Firat, Schrittwieser, et~al.]{gemini1_5}
Machel Reid, Nikolay Savinov, Denis Teplyashin, Dmitry Lepikhin, Timothy Lillicrap, Jean-baptiste Alayrac, Radu Soricut, Angeliki Lazaridou, Orhan Firat, Julian Schrittwieser, et~al.
\newblock Gemini 1.5: Unlocking multimodal understanding across millions of tokens of context.
\newblock \emph{arXiv preprint arXiv:2403.05530}, 2024.

\bibitem[Saad-Falcon et~al.(2024)Saad-Falcon, Lafuente, Natarajan, Maru, Todorov, Guha, Buchanan, Chen, Guha, R{\'e}, et~al.]{saad2024archon}
Jon Saad-Falcon, Adrian~Gamarra Lafuente, Shlok Natarajan, Nahum Maru, Hristo Todorov, Etash Guha, E~Kelly Buchanan, Mayee Chen, Neel Guha, Christopher R{\'e}, et~al.
\newblock Archon: An architecture search framework for inference-time techniques.
\newblock \emph{arXiv preprint arXiv:2409.15254}, 2024.

\bibitem[Shang et~al.(2024)Shang, Li, Zhao, Ma, Liu, Xu, and Li]{Agentsquare}
Yu~Shang, Yu~Li, Keyu Zhao, Likai Ma, Jiahe Liu, Fengli Xu, and Yong Li.
\newblock Agentsquare: Automatic llm agent search in modular design space.
\newblock \emph{arXiv preprint arXiv:2410.06153}, 2024.

\bibitem[Team(2024{\natexlab{a}})]{claude}
Anthropic Team.
\newblock The claude 3 model family: Opus, sonnet, haiku.
\newblock \emph{https://www-cdn.anthropic.com/de8ba9b01c9ab7cbabf5c33b80b7bbc618857627/Model\_Card\_Claude\_3.pdf}, 2024{\natexlab{a}}.

\bibitem[Team(2025{\natexlab{a}})]{deepseek_r1}
Deepseek Team.
\newblock Deepseek-r1: Incentivizing reasoning capability in llms via reinforcement learning.
\newblock \emph{arXiv preprint arXiv:2501.12948}, 2025{\natexlab{a}}.

\bibitem[Team(2025{\natexlab{b}})]{kimi_k1_5}
Kimi Team.
\newblock Kimi k1.5: Scaling reinforcement learning with llms.
\newblock \emph{arXiv preprint 2501.12599v1}, 2025{\natexlab{b}}.

\bibitem[Team(2024{\natexlab{b}})]{qvq}
Qwen Team.
\newblock Qvq: To see the world with wisdom.
\newblock \emph{https://qwenlm.github.io/blog/qvq-72b-preview/}, 2024{\natexlab{b}}.

\bibitem[Team(2024{\natexlab{c}})]{qwq}
Qwen Team.
\newblock Qwq: Reflect deeply on the boundaries of the unknown.
\newblock \emph{https://qwenlm.github.io/blog/qwq-32b-preview/}, 2024{\natexlab{c}}.

\bibitem[Wang et~al.(2023)Wang, Xu, Lan, Hu, Lan, Lee, and Lim]{wang2023plan}
Lei Wang, Wanyu Xu, Yihuai Lan, Zhiqiang Hu, Yunshi Lan, Roy Ka-Wei Lee, and Ee-Peng Lim.
\newblock Plan-and-solve prompting: Improving zero-shot chain-of-thought reasoning by large language models.
\newblock \emph{arXiv preprint arXiv:2305.04091}, 2023.

\bibitem[Wang et~al.(2025)Wang, Yang, Li, Wang, and Aragam]{wang2025scoreflow}
Yinjie Wang, Ling Yang, Guohao Li, Mengdi Wang, and Bryon Aragam.
\newblock Scoreflow: Mastering llm agent workflows via score-based preference optimization.
\newblock \emph{arXiv preprint arXiv:2502.04306}, 2025.

\bibitem[Wei et~al.(2022)Wei, Wang, Schuurmans, Bosma, Xia, Chi, Le, Zhou, et~al.]{CoT}
Jason Wei, Xuezhi Wang, Dale Schuurmans, Maarten Bosma, Fei Xia, Ed~Chi, Quoc~V Le, Denny Zhou, et~al.
\newblock Chain-of-thought prompting elicits reasoning in large language models.
\newblock \emph{Advances in neural information processing systems}, 35:\penalty0 24824--24837, 2022.

\bibitem[Wu et~al.(2023)Wu, Bansal, Zhang, Wu, Li, Zhu, Jiang, Zhang, Zhang, Liu, et~al.]{autogen}
Qingyun Wu, Gagan Bansal, Jieyu Zhang, Yiran Wu, Beibin Li, Erkang Zhu, Li~Jiang, Xiaoyun Zhang, Shaokun Zhang, Jiale Liu, et~al.
\newblock Autogen: Enabling next-gen llm applications via multi-agent conversation.
\newblock \emph{arXiv preprint arXiv:2308.08155}, 2023.

\bibitem[Yang et~al.(2024)Yang, Yang, Zhang, Hui, Zheng, Yu, Li, Liu, Huang, Wei, et~al.]{qwen2_5}
An~Yang, Baosong Yang, Beichen Zhang, Binyuan Hui, Bo~Zheng, Bowen Yu, Chengyuan Li, Dayiheng Liu, Fei Huang, Haoran Wei, et~al.
\newblock Qwen2. 5 technical report.
\newblock \emph{arXiv preprint arXiv:2412.15115}, 2024.

\bibitem[Yang et~al.(2023)Yang, Wang, Lu, Liu, Le, Zhou, and Chen]{yang2023large}
Chengrun Yang, Xuezhi Wang, Yifeng Lu, Hanxiao Liu, Quoc~V Le, Denny Zhou, and Xinyun Chen.
\newblock Large language models as optimizers.
\newblock \emph{arXiv preprint arXiv:2309.03409}, 2023.

\bibitem[Yuan et~al.(2024)Yuan, Song, Chen, Tan, Li, and Yang]{yuan2024evoagent}
Siyu Yuan, Kaitao Song, Jiangjie Chen, Xu~Tan, Dongsheng Li, and Deqing Yang.
\newblock Evoagent: Towards automatic multi-agent generation via evolutionary algorithms.
\newblock \emph{arXiv preprint arXiv:2406.14228}, 2024.

\bibitem[Yuksekgonul et~al.(2024)Yuksekgonul, Bianchi, Boen, Liu, Huang, Guestrin, and Zou]{yuksekgonul2024textgrad}
Mert Yuksekgonul, Federico Bianchi, Joseph Boen, Sheng Liu, Zhi Huang, Carlos Guestrin, and James Zou.
\newblock Textgrad: Automatic" differentiation" via text.
\newblock \emph{arXiv preprint arXiv:2406.07496}, 2024.

\bibitem[Zhang et~al.(2024{\natexlab{a}})Zhang, Yue, Sun, Wan, Yu, Fang, Wang, and Cheng]{g_designer}
Guibin Zhang, Yanwei Yue, Xiangguo Sun, Guancheng Wan, Miao Yu, Junfeng Fang, Kun Wang, and Dawei Cheng.
\newblock G-designer: Architecting multi-agent communication topologies via graph neural networks.
\newblock \emph{arXiv preprint arXiv:2410.11782}, 2024{\natexlab{a}}.

\bibitem[Zhang et~al.(2025)Zhang, Niu, Fang, Wang, Bai, and Wang]{MaAS}
Guibin Zhang, Luyang Niu, Junfeng Fang, Kun Wang, Lei Bai, and Xiang Wang.
\newblock Multi-agent architecture search via agentic supernet.
\newblock \emph{arXiv preprint arXiv:2502.04180}, 2025.

\bibitem[Zhang et~al.(2024{\natexlab{b}})Zhang, Xiang, Yu, Teng, Chen, Chen, Zhuge, Cheng, Hong, Wang, et~al.]{AFLOW}
Jiayi Zhang, Jinyu Xiang, Zhaoyang Yu, Fengwei Teng, Xionghui Chen, Jiaqi Chen, Mingchen Zhuge, Xin Cheng, Sirui Hong, Jinlin Wang, et~al.
\newblock Aflow: Automating agentic workflow generation.
\newblock \emph{arXiv preprint arXiv:2410.10762}, 2024{\natexlab{b}}.

\bibitem[Zhang et~al.(2024{\natexlab{c}})Zhang, Zhang, Liu, Song, Wang, Krishna, and Wu]{zhang2024offline}
Shaokun Zhang, Jieyu Zhang, Jiale Liu, Linxin Song, Chi Wang, Ranjay Krishna, and Qingyun Wu.
\newblock Offline training of language model agents with functions as learnable weights.
\newblock In \emph{Forty-first International Conference on Machine Learning}, 2024{\natexlab{c}}.

\bibitem[Zheng et~al.(2024)Zheng, Zhang, Zhang, Ye, Luo, Feng, and Ma]{llamafactory}
Yaowei Zheng, Richong Zhang, Junhao Zhang, Yanhan Ye, Zheyan Luo, Zhangchi Feng, and Yongqiang Ma.
\newblock Llamafactory: Unified efficient fine-tuning of 100+ language models.
\newblock \emph{arXiv preprint arXiv:2403.13372}, 2024.

\bibitem[Zhou et~al.(2024{\natexlab{a}})Zhou, Pujara, Ren, Chen, Cheng, Le, Chi, Zhou, Mishra, and Zheng]{zhou2024self}
Pei Zhou, Jay Pujara, Xiang Ren, Xinyun Chen, Heng-Tze Cheng, Quoc~V Le, Ed~Chi, Denny Zhou, Swaroop Mishra, and Huaixiu~Steven Zheng.
\newblock Self-discover: Large language models self-compose reasoning structures.
\newblock \emph{Advances in Neural Information Processing Systems}, 37:\penalty0 126032--126058, 2024{\natexlab{a}}.

\bibitem[Zhou et~al.(2024{\natexlab{b}})Zhou, Ou, Ding, Li, Wu, Wang, Chen, Wang, Xu, Zhang, et~al.]{zhou2024symbolic}
Wangchunshu Zhou, Yixin Ou, Shengwei Ding, Long Li, Jialong Wu, Tiannan Wang, Jiamin Chen, Shuai Wang, Xiaohua Xu, Ningyu Zhang, et~al.
\newblock Symbolic learning enables self-evolving agents.
\newblock \emph{arXiv preprint arXiv:2406.18532}, 2024{\natexlab{b}}.

\bibitem[Zhuge et~al.(2023)Zhuge, Liu, Faccio, Ashley, Csord{\'a}s, Gopalakrishnan, Hamdi, Hammoud, Herrmann, Irie, et~al.]{zhuge2023mindstorms}
Mingchen Zhuge, Haozhe Liu, Francesco Faccio, Dylan~R Ashley, R{\'o}bert Csord{\'a}s, Anand Gopalakrishnan, Abdullah Hamdi, Hasan Abed Al~Kader Hammoud, Vincent Herrmann, Kazuki Irie, et~al.
\newblock Mindstorms in natural language-based societies of mind.
\newblock \emph{arXiv preprint arXiv:2305.17066}, 2023.

\bibitem[Zhuge et~al.(2024)Zhuge, Wang, Kirsch, Faccio, Khizbullin, and Schmidhuber]{GPTSwarm}
Mingchen Zhuge, Wenyi Wang, Louis Kirsch, Francesco Faccio, Dmitrii Khizbullin, and J{\"u}rgen Schmidhuber.
\newblock Gptswarm: Language agents as optimizable graphs.
\newblock In \emph{Forty-first International Conference on Machine Learning}, 2024.

\bibitem[Zhuo et~al.(2024)Zhuo, Vu, Chim, Hu, Yu, Widyasari, Yusuf, Zhan, He, Paul, et~al.]{bigcodebench}
Terry~Yue Zhuo, Minh~Chien Vu, Jenny Chim, Han Hu, Wenhao Yu, Ratnadira Widyasari, Imam Nur~Bani Yusuf, Haolan Zhan, Junda He, Indraneil Paul, et~al.
\newblock Bigcodebench: Benchmarking code generation with diverse function calls and complex instructions.
\newblock \emph{arXiv preprint arXiv:2406.15877}, 2024.

\end{thebibliography}
\bibliographystyle{plainnat}

\clearpage
\appendix
\section{More Implementation Details}
\label{details}
In the workflow optimization process of our method, we evaluate each workflow for 3 times as external execution feedback. We use the following 6 operators based on Aflow~\citep{AFLOW} as our base operators:
\begin{itemize}
\item Code Generator: generates code solutions for a given problem.
\item Format Generator: produces formatted answers for a given problem.
\item Ensemble Operator: combines multiple solutions or approaches to create a more robust final result.
\item Review Operator: evaluates solutions for correctness, efficiency, and adherence to requirements.
\item Revise Operator: refines solutions based on feedback from the review process.
\item Code Test Operator: executes and validates code solutions against test cases to ensure functionality.
\end{itemize}

 For our supervised fine-tuning (SFT), we utilized approximately 1,400 items sourced from three datasets generated by R1 through our optimization process. We conducted the training using LLaMA-Factory~\citep{llamafactory}, with a per-device training batch size of 1, gradient accumulation over 2 steps, a learning rate of 1e-5, and max train epochs of 3. For the reinforcement learning (RL) phase, we set the scaling factor $k$ to 1.1, the threshold $T$ to 3, rollout number $m$ to 5, and max episodes to 5.

\section{More Cases of Workflow}
In this section, we present additional cases of both successful and failed results generated by \ourmethod.
\label{case_more}
\subsection{Successful Cases}
Fig.~\ref{fig:python-code-1}, Fig.~\ref{fig:python-code-2} and Fig.~\ref{fig:python-code-3} demonstrate three successful cases generated by \ourmethod for BigCodeBench, HumanEval and MBPP.
\begin{figure}[tp]
    \centering
    \begin{lstlisting}[language=Python, basicstyle=\scriptsize\ttfamily]
async def __call__(self, problem: str, test_cases: list):
    # Extract key problem requirements and constraints
    problem_analysis = await self.custom(
        input=problem,
        instruction=prompt_custom.GRAPH_ANALYZE_PROMPT
    )
    
    # Generate initial solution with problem analysis context
    solution = await self.custom_code_generate(
        problem=problem, 
        entry_point=entry_point, 
        instruction=f"{prompt_custom.GRAPH_GENERATE_PROMPT}\nProblem Analysis: {problem_analysis['response']}"
    )
    
    current_solution = solution['response']
    
    # Loop for max_iterations to improve the solution if tests fail
    for iteration in range(3):
        # Test current solution
        test_results = await self.custom(
            input=f"Solution: {current_solution}\nTest Cases: {test_cases}", 
            instruction=prompt_custom.GRAPH_TEST_PROMPT
        )
        
        # If tests pass, return the solution
        if "failed" not in test_results['response'].lower():
            return current_solution, self.llm.cost_manager.total_cost
        
        # Improve solution with test feedback and problem analysis
        improved_solution = await self.custom_code_generate(
            problem=problem,
            entry_point=entry_point,
            instruction=f"{prompt_custom.GRAPH_IMPROVE_PROMPT}\nProblem Analysis: {problem_analysis['response']}\nCurrent Solution: {current_solution}\nTest Results: {test_results['response']}\nIteration: {iteration + 1}/{max_iterations}"
        )
        
        # Update current solution
        current_solution = improved_solution['response']
    
    # Return the last solution attempt
    return current_solution, self.llm.cost_manager.total_cost
    \end{lstlisting}
    \caption{Successful workflow generated by \ourmethod-14B for the BigCodeBench task of \emph{Generating and plot weather data for a specified date range.}}
    \label{fig:python-code-1}
\end{figure}

\begin{figure}[tp]
    \centering
    \begin{lstlisting}[language=Python, basicstyle=\scriptsize\ttfamily]
async def __call__(self, problem: str, test_cases: list):
    # Generate solution
    solution = await self.custom_code_generate(problem=problem, entry_point=entry_point, instruction=prompt_custom.GRAPH_GENERATE_PROMPT)
    
    # Test solution
    test_results = await self.custom(
        input=f"Solution: {solution['response']}\nTest Cases: {test_cases}", 
        instruction=prompt_custom.GRAPH_TEST_PROMPT
    )
    
    # Optimize solution if tests fail
    if "failed" in test_results['response'].lower():
        improved_solution = await self.custom_code_generate(
            problem=problem,
            instruction=f"{prompt_custom.GRAPH_IMPROVE_PROMPT}\n\nTest Results: {test_results['response']}"
        )
        return improved_solution['response'], self.llm.cost_manager.total_cost
    
    return solution['response'], self.llm.cost_manager.total_cost
    \end{lstlisting}
    \caption{Successful workflow generated by \ourmethod-14B for the MBPP task of \emph{Writing a function to reverse words in a given string.}}
    \label{fig:python-code-3}
\end{figure}
\begin{figure}[tp]
    \centering
    \begin{lstlisting}[language=Python, basicstyle=\scriptsize\ttfamily]
async def __call__(self, problem: str, test_cases: list):
    # 1. Prepare generation prompts
    gen_prompts = [
        prompt_custom.CODE_GENERATE_PROMPT_A,
        prompt_custom.CODE_GENERATE_PROMPT_B,
        prompt_custom.CODE_GENERATE_PROMPT_C,
    ]
    
    # 2. Generate three candidate solutions
    sols = [
        (await self.custom_code_generate(
            problem=problem,
            instruction=inst
        ))['response'] for inst in gen_prompts
    ]
    
    # 3. Ensemble: pick or merge the best solution
    ensemble_input = "Solutions:\n" + "\n---\n".join(sols)
    
    best = (await self.custom(
        input=ensemble_input,
        instruction=prompt_custom.ENSEMBLE_PROMPT
    ))['response']
    
    # 4. Return the best solution and total cost
    return best, self.llm.cost_manager.total_cost
    \end{lstlisting}
    \caption{Successful workflow generated by \ourmethod-14B for the HumanEval task of \emph{Returning list of prime factors of given integer in the order from smallest to largest.}}
    \label{fig:python-code-2}
\end{figure}

\subsection{Failure Cases}
Fig.~\ref{fig:python-code2-1}, Fig.~\ref{fig:python-code2-2} and Fig.~\ref{fig:python-code2-3} demonstrate three failure cases generated by \ourmethod for BigCodeBench, HumanEval and MBPP.
\begin{figure}[t]
    \centering
    \begin{lstlisting}[language=Python, basicstyle=\scriptsize\ttfamily]
async def __call__(self, problem: str, test_cases: list):
    # 1. Analyze the problem
    analysis = await self.custom(
        input=problem,
        instruction=prompt_custom.ANALYZE_PROMPT
    )
    
    # 2. Generate initial solution
    solution = await self.custom_code_generate(
        problem=problem,
        instruction=prompt_custom.GENERATE_PROMPT,
        context=analysis['response']
    )
    
    # 3. Test with a subset of test cases
    sample_tests = test_cases[:min(3, len(test_cases))]
    test_result = await self.custom(
        input=f"Solution:\n{solution['response']}\nTests:\n{sample_tests}",
        instruction=prompt_custom.TEST_PROMPT
    )
    
    # 4. Handle based on test results
    if "all passed" in test_result['response'].lower():
        # Optimize if all tests pass
        final = await self.custom_code_generate(
            problem=problem,
            instruction=prompt_custom.OPTIMIZE_PROMPT,
            context=f"Solution:\n{solution['response']}"
        )
        return final['response'], self.llm.cost_manager.total_cost
    else:
        # Fix issues if tests fail
        fixed = await self.custom_code_generate(
            problem=problem,
            instruction=prompt_custom.FIX_PROMPT,
            context=f"Solution:\n{solution['response']}\nTest results:\n{test_result['response']}"
        )
        return fixed['response'], self.llm.cost_manager.total_cost
    \end{lstlisting}
    \caption{Failed workflow generated by \ourmethod-14B for the MBPP task of \emph{Writing a function to count those characters which have vowels as their neighbors in the given string.}}
    \label{fig:python-code2-3}
\end{figure}
\begin{figure}[ht]
    \centering
    \begin{lstlisting}[language=Python, basicstyle=\scriptsize\ttfamily]
async def __call__(self, problem: str, test_cases: list):
    async def __call__(self, problem: str, test_cases: list):
    # Extract key problem requirements and constraints with detailed specifications
    problem_analysis = await self.custom(
        input=problem,
        instruction=prompt_custom.GRAPH_ANALYZE_PROMPT
    )
    
    # Initial approach planning with appropriate algorithm selection
    algorithm_selection = await self.custom(
        input=f"Problem: {problem}\nAnalysis: {problem_analysis['response']}",
        instruction=prompt_custom.GRAPH_ALGORITHM_SELECTION_PROMPT
    )
    
    # Generate solution with problem analysis and algorithm selection
    solution = await self.custom_code_generate(
        problem=problem, 
        entry_point=entry_point, 
        instruction=f"{prompt_custom.GRAPH_GENERATE_PROMPT}\nProblem Analysis: {problem_analysis['response']}\nAlgorithm Selection: {algorithm_selection['response']}"
    )
    
    # Self-review for logical errors and optimization opportunities
    code_review = await self.custom(
        input=f"Code: {solution['response']}",
        instruction=prompt_custom.GRAPH_CODE_REVIEW_PROMPT
    )
    
    # Generate improved solution with comprehensive context
    improved_solution = await self.custom_code_generate(
        problem=problem,
        entry_point=entry_point,
        instruction=f"{prompt_custom.GRAPH_IMPROVE_PROMPT}\nProblem Analysis: {problem_analysis['response']}\nAlgorithm Selection: {algorithm_selection['response']}\nCode Review: {code_review['response']}"
    )
    
    # Add explanation to the solution
    solution_explanation = await self.custom(
        input=f"Solution: {improved_solution['response']}\nProblem: {problem}",
        instruction=prompt_custom.GRAPH_SOLUTION_EXPLANATION_PROMPT
    )
    
    final_solution = f"# Solution\n{improved_solution['response']}\n\n# Explanation\n{solution_explanation['response']}"
    return final_solution, self.llm.cost_manager.total_cost

    \end{lstlisting}
    \caption{Failed workflow generated by \ourmethod-14B for the BigCodeBench task of \emph{Extracting the text and href attributes of all anchor tags from a given URL's HTML content.}}
    \label{fig:python-code2-1}
\end{figure}

\begin{figure}[tp]
    \centering
    \begin{lstlisting}[language=Python, basicstyle=\scriptsize\ttfamily,showspaces=false, showtabs=false]
async def __call__(self, problem: str, test_cases: list):
    # 1. Generate solution
    solution = (await self.custom_code_generate(
        problem=problem,
        instruction=prompt_custom.CODE_GENERATE_PROMPT
    ))['response']
    
    # 2. Test and refine solution
    refined_solution = (await self.custom_code_generate(
        problem=problem,
        instruction=prompt_custom.REFINE_PROMPT,
        context=f"Initial solution:\n{solution}\n\nTest cases:\n{test_cases}"
    ))['response']
    
    # 3. Return the solution and total cost
    return refined_solution, self.llm.cost_manager.total_cost
    \end{lstlisting}
    \caption{Failed workflow generated by \ourmethod-14B for the HumanEval task of \emph{Evaluating whether the given number n can be written as the sum of exactly 4 positive even numbers.}}
    \label{fig:python-code2-2}
\end{figure}

\end{document}